\documentclass[10pt, conference]{IEEEtran}

\usepackage{amsmath,amssymb,amsfonts}
\usepackage{graphicx}
\usepackage{textcomp}

\usepackage{amsthm}
\usepackage{xcolor}
\usepackage[figuresright]{rotating}

\usepackage{graphicx}
\graphicspath{{/}{fig/}}

\usepackage{array}
\usepackage{textcomp}
\usepackage{xcolor}
\usepackage{multirow}
\usepackage{booktabs}
\usepackage{enumitem}

\usepackage{mathtools}
\usepackage{breqn}
\usepackage{float}

\usepackage{caption}
\usepackage{subcaption}

\usepackage{mathtools}
\usepackage{float}

\usepackage{pgfplots}
\pgfplotsset{compat=1.7}
\usepgfplotslibrary{groupplots}

\usepackage{multirow,tabularx}

\usepackage{pifont}
\newcommand{\cmark}{\ding{51}}%
\newcommand{\xmark}{\ding{55}}%

\newlength\figureheight
\newlength\figurewidth
\title{
    Sim-to-Real Transfer in Deep Reinforcement Learning for Robotics: a Survey
}

\author{
    \IEEEauthorblockN{
        Wenshuai Zhao\textsuperscript{1},
        Jorge Peña Queralta\textsuperscript{1},
        Tomi Westerlund\textsuperscript{1}
    }\\[6pt]
    \IEEEauthorblockA{
        \textsuperscript{1}{Turku Intelligent Embedded and Robotic Systems Lab, University of Turku, Finland} \\
        Emails: \textsuperscript{1}\{wezhao, jopequ, tovewe\}@utu.fi
    }
}

\begin{document}


\maketitle
\IEEEpubidadjcol

\begin{abstract}

    Deep reinforcement learning has recently seen huge success across multiple areas in the robotics domain. Owing to the limitations of gathering real-world data, i.e., sample inefficiency and the cost of collecting it, simulation environments are utilized for training the different agents. This not only aids in providing a potentially infinite data source, but also alleviates safety concerns with real robots. Nonetheless, the gap between the simulated and real worlds degrades the performance of the policies once the models are transferred into real robots. Multiple research efforts are therefore now being directed towards closing this sim-to-real gap and accomplish more efficient policy transfer. Recent years have seen the emergence of multiple methods applicable to different domains, but there is a lack, to the best of our knowledge, of a comprehensive review summarizing and putting into context the different methods. In this survey paper, we cover the fundamental background behind sim-to-real transfer in deep reinforcement learning and overview the main methods being utilized at the moment: domain randomization, domain adaptation, imitation learning, meta-learning and knowledge distillation. We categorize some of the most relevant recent works, and outline the main application scenarios. Finally, we discuss the main opportunities and challenges of the different approaches and point to the most promising directions.
    

\end{abstract}

\begin{IEEEkeywords}
    Deep Reinforcement Learning; Robotics; Sim-to-Real; Transfer Learning; Meta Learning; Domain Randomization; Knowledge Distillation; Imitation Learning;
\end{IEEEkeywords}

\IEEEpeerreviewmaketitle
\section{Introduction}

\begin{figure}
    \centering
    \includegraphics[width=0.48\textwidth]{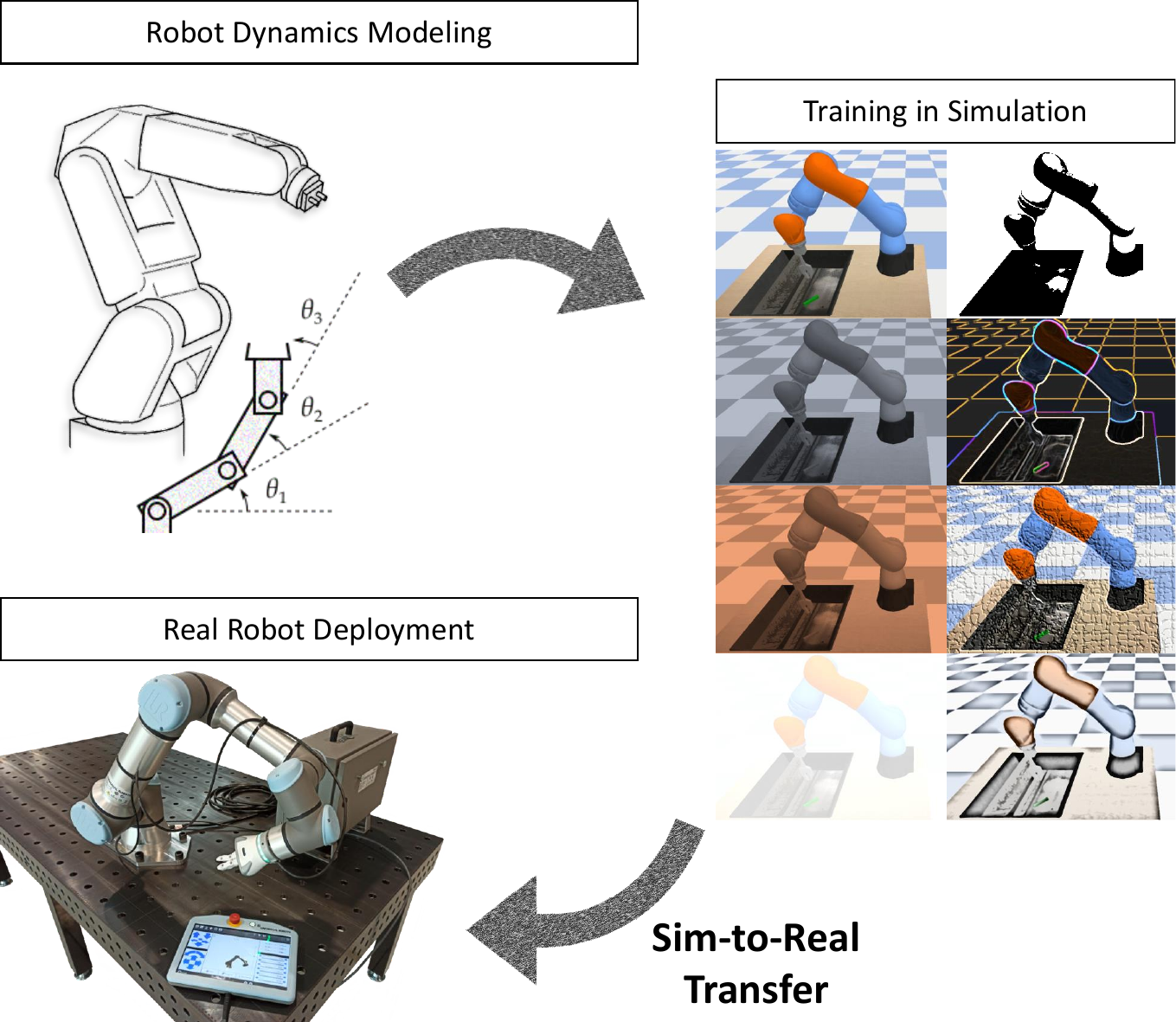}
    \vspace{1.75ex}
    \caption{Conceptual view of a simulation-to-reality transfer process. One of the most common methods is domain randomization, through which different parameters of the simulator (e..g, colors, textures, dynamics) are randomized to produce more robust policies.}
    \label{fig:concept}
    \vspace{-0.5em}
\end{figure}

Reinforcement learning (RL) algorithms have been increasingly adopted by the robotics community over the past years to control complex robots or multi-robot systems~\cite{arulkumaran2017brief, nguyen2020deep}, or provide end-to-end policies from perception to control~\cite{cheng2019end}. Inspired by the way we learn through trial-and-error processes, RL algorithms base their knowledge acquisition in the rewards that agents obtain when they act in certain manners given different experiences. This naturally requires a large number of episodes, and therefore the learning limitations in terms of time and experience variability in real-world scenarios is evident. Moreover, learning with real robots requires the consideration of potentially dangerous or unexpected behaviors in safety-critical applications~\cite{garcia2015comprehensive}. Deep reinforcement learning (DRL) algorithms have been successfully deployed in various types of simulation environments, yet their success beyond simulated worlds has been limited. An exception to this is, however, robotic tasks involving object manipulation~\cite{rajeswaran2017learning, matas2018sim}. In this survey, we review the most relevant works that try to answer a key research question in this direction: how to exploit simulation-based training in real-world settings by transferring the knowledge and adapting the policies accordingly (Fig.~\ref{fig:concept}).

Simulation-based training provides data at low-cost, but involves inherent mismatches with real-world settings. Bridging the gap between simulation and reality requires, first of all, methods that are able to account for mismatches in both sensing and actuation. The former aspect has been widely studied in recent years within the deep learning field, for instance with adversarial attacks on computer vision algorithms~\cite{akhtar2018threat}. The latter risk can be minimized through more realistic simulation. In both of these cases, some of the current approaches include works that introduce perturbances in the environment~\cite{wenshuai2020sim2real} or focus on domain randomization~\cite{muratore2020bayesian}. Another key aspect to take into account is that an agent deployed in the real world will potentially be exposed to novel experiences that were not present in the simulations~\cite{ramakrishnan2020blind}, as well as the potential need to adapt their policies to encompass wider sets of tasks. Some of the approaches to bridge the gap in this direction rely on meta learning~\cite{arndt2019meta} or continual learning~\cite{traore2019continual}, among others.

The methods described above focus on extracting knowledge from simulation-trained agents in order to deploy them in real-life scenarios. However, other approaches exist to the same end. In recent years, simulators have been progressing towards more realistic scenarios and physics engines: Airsim~\cite{shah2018airsim}, CARLA~\cite{dosovitskiy2017carla}, RotorS~\cite{furrer2016rotors, mccord2019distributed}, and others~\cite{todorov2012mujoco}. With some of these simulators, part of the aim is to be able to deploy the robotic agents directly into the real world by providing training data and experiences with minimal mismatches between real and simulated settings. Other research efforts have been directed towards increasing safety during training in real-settings. Safety is one of the main challenges towards achieving online training of complex agents in the real-world, from robot arms to self-driving cars~\cite{garcia2015comprehensive}. In this direction, recent works have shown promising results towards safe DRL that is able to ensure convergence even while reducing the exploration space~\cite{cheng2019end}. In this survey, we do not cover specific simulators or techniques for direct learning in real-world settings, but instead focus on describing the main methods for transferring knowledge learned in simulation towards their deployment in real robotic platforms.

This is, to the best of our knowledge, the first survey that describes the different methods being utilized towards closing the simulation-to-reality gap in DRL for robotics. We also concentrate on describing the main application fields of current research efforts. We discuss recent works from a wider point of view by including related research directions in the areas of transfer learning and domain adaptation, knowledge distillation, and meta reinforcement learning. While other surveys have focused on transfer learning techniques~\cite{zhuang2020comprehensive} or safe reinforcement learning~\cite{garcia2015comprehensive}, we provide a different point of view with an emphasis on DRL policy transfer in the robotics domain. Finally, there is also a significant amount of publications deploying DRL policies on real robots. In this survey, nonetheless, we focus on those works that specifically tackle issues in sim-to-real transfer. The focus is mostly in end-to-end approaches, but we also describe relevant research where sim-to-real transfer techniques are applied to the sensing aspects of robotic operation, primarily the transfer of DL vision algorithms to real robots.

The rest of this paper is organized as follows. In Section~\ref{sec:background}, we briefly introduce the main approaches to DRL, together with related research directions in knowledge distillation, transfer, adaptation and meta learning. Section~\ref{sec:methods} then delves into the different approaches being taken towards closing the simulation-to-reality gap, with Section~\ref{sec:applications} focusing on the most relevant application areas. Then, we discuss open challenges and promising research directions in Section~\ref{sec:discusion}. Finally, Section~\ref{sec:conclusion} concludes this survey.


\section{Background}
\label{sec:background}

\begin{figure}
        \centering
        \includegraphics[width=0.49\textwidth]{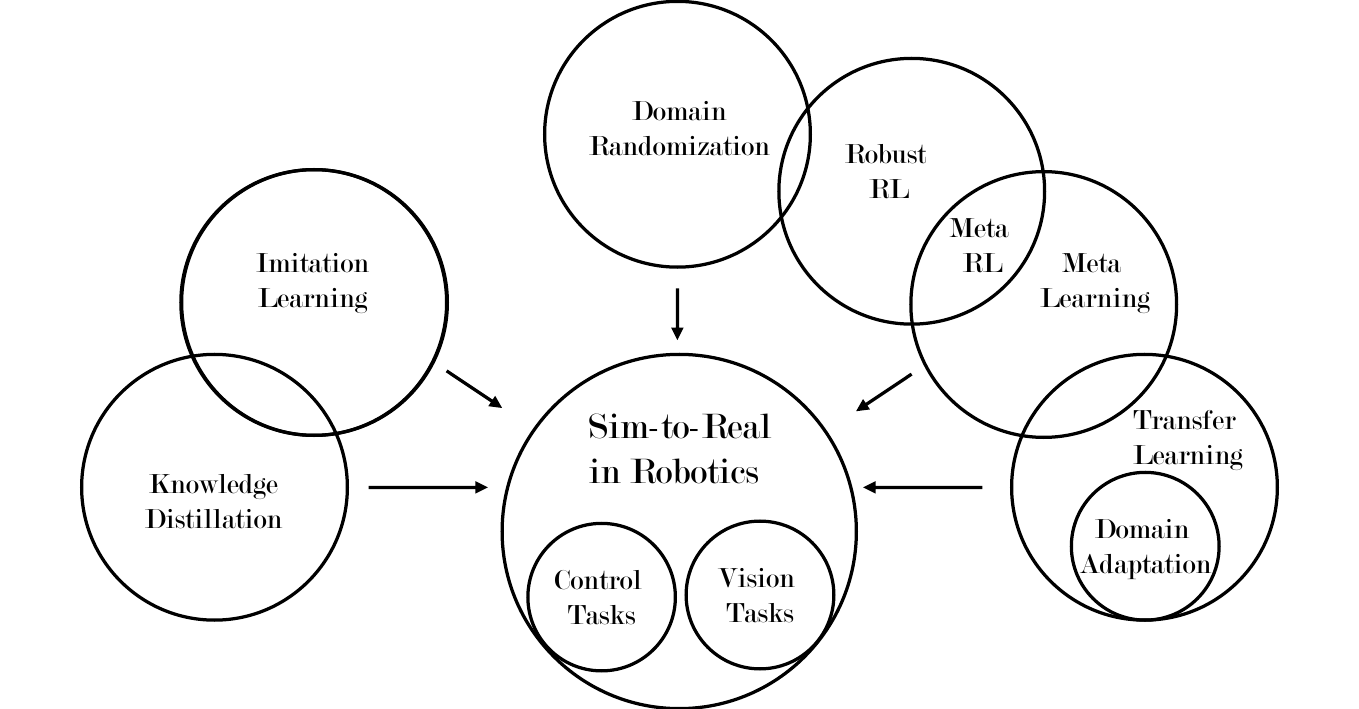}
        \caption{Illustration of the different methods related to sim-to-real transfer in deep reinforcement learning and their relationships.} 
        \vspace{-1em}
        \label{fig:RC}
\end{figure}

Sim-to-real is a very comprehensive concept and applied in many fields including robotics and classic machine vision tasks. Thereby quite a few methods and concepts intersect with this aim including transfer learning, robust RL, and meta learning. In this section, we briefly introduce the concepts of deep reinforcement learning, knowledge distillation, transfer learning and domain adaption, before going into more details about sim-to-real transfer methods for DRL. The relationship between there concepts is illustrated in Fig.~\ref{fig:RC}.

\subsection{Deep Reinforcement Learning}

A standard reinforcement learning (RL) task can be regarded as a sequential decision making setup which consists of an \textit{agent} interacting with an \textit{environment} in discrete steps. The \textit{agent} takes an action $a_{t}$ at each timestep \textit{t}, causing the \textit{environment} to change its state from $s_{t}$ to $s_{t+1}$ with a transition probability $p(s_{t+1}\vert s_{t},a_{t})$. This setup can be regarded as a Markov decision process (MDP) with a set of states $s\in \mathcal{S}$, actions $a\in \mathcal{A}$, transitions $p\in \mathcal{P}$ and rewards $r\in \mathcal{R}$. Therefore we can define this MDP as a tuple \eqref{eq:MDP}.
\begin{equation}
    D \equiv (\mathcal{S}, \mathcal{A}, \mathcal{P}, \mathcal{R})
    \label{eq:MDP}
\end{equation} 

The objective of reinforcement learning is to maximize the expected reward by choosing an optimal policy which will be represented via a deep neural network in DRL. Accelerated by modern computation capacity, DRL has shown significant success on various applications~\cite{arulkumaran2017brief, queralta2020collaborative}, but particular in the simulated environment~\cite{berner2019dota}. Therefore, how to transfer this success from simulation to reality is drawing more and more attention, which is also the motivation of this paper.

\subsection{Sim-to-Real Transfer}

Transferring DRL policies form simulation environments to reality is a necessary step towards more complex robotic systems that have DL-defined controllers. This, however, is not a problem specific to DRL algorithms, but ML in general. While most DRL algorithms provide end-to-end policies, i.e., control mechanisms that take raw sensor data as inputs and produce direct actuation commands as outputs, these two dimensions of robotics can be separated. Closing the gap between simulation and reality gap in terms of actuation requires simulators to be more accurate, and to account for variability in agent dynamics. On the sensing part, however, the problem can be considered wider, as it also involves the more general ML problem of facing situations in the real world that have not appeared in simulation~\cite{ramakrishnan2020blind}.  In this paper, we focus mostly on end-to-end models, and overview both research directed towards system modeling and dynamics randomization, as well as research introducing randomization from the sensing point of view.

\subsection{Transfer Learning and Domain Adaptation}

Transfer learning aims at improving the performance of target learners on target domains by transferring the knowledge contained in different but related source domains~\cite{zhuang2020comprehensive}. In this way, transfer learning can reduce the dependence of target domain data when constructing target learners. 

Domain adaptation is a subset of transfer learning methods. It specifies the situation when we have sufficient source domain labeled data and the same single task as the target task, but without or very few target domain data. In sim-to-real robotics, researchers tend to employ a simulator to train the RL model and then deploy it in the realistic environment, where we should take advantage of the domain adaptation techniques in order to transfer the simulation based model well.  

\subsection{Knowledge Distillation}

Large networks are typical in DRL with high-dimensional input data (e.g, complex visual tasks). Policy distillation is the process of extracting knowledge to train a new network that is able to maintain a similarly expert level while being significantly smaller and more efficient~\cite{rusu2015policy}. In these set-ups, the two networks are typically called \textit{teacher} and \textit{student}. The student is trained in a supervised manner with data generated by the teacher network. In~\cite{traore2019continual}, the authors presented DisCoRL, a modular, effective and scalable pipeline for continual DRL. DisCoRL has been succesfully applied to multiple tasks learned by different teachers, with their knowledge being distilled to a single student network.



\subsection{Meta Reinforcement Learning}

Meta Learning, namely learning to learn, aims to learn the adaptation ability to unseen test tasks from multiple training tasks. A good meta learning model should be trained across a variety of learning tasks and optimized for the best performance over a distribution of tasks, including potentially unseen tasks when tested. This spirit can be applied on both supervised learning and reinforcement learning, and in the latter case it is called meta reinforcement learning (MetaRL)~\cite{wang2016learning}. 

The overall configuration of MetaRL is similar to an ordinary RL algorithm, except that MetaRL usually implements an LSTM policy and incorporates the last reward $r_{t-1}$ and last action $a_{t-1}$ into the current policy observation. In this case, the LSTM's hidden states serve as a memory for tracking characteristics of the trajectories. Therefore, MetaRL draws knowledge from past training. 





\subsection{Robust RL and Imitation Learning}


Robust RL~\cite{morimoto2005robust} was proposed quite early as a new RL paradigm that explicitly takes into account input disturbances as well as modeling errors. It considers a bad, or even adversarial model and tries to maximize the reward as a optimization problem~\cite{tessler2019action, mankowitz2019robust}. 

Imitation learning proposes to employ expert demonstration or trajectories instead of manually constructing a fixed reward function to train RL agents. The methods of imitation learning can be broadly classified into two key areas:~\textit{behaviour cloning} where an agent learns a mapping
from observations to actions given demonstrations~\cite{pomerleau1989alvinn, ross2011reduction} and ~\textit{inverse reinforcement learning} where an agent
attempts to estimate a reward function that describes the
given demonstrations~\cite{ng2000algorithms}. Because it aims to give a robust reward for RL agents, sometimes imitation learning can be utilized to obtain robust RL or sim-to-real transfer~\cite{yan2017sim}.


\begin{sidewaystable*}
    \centering
    \caption{\footnotesize{Classification of the most relevant publications in Sim2Real Transfer.}}
    \label{tab:papers}
    \footnotesize
    \renewcommand{\arraystretch}{1.42}
    \begin{tabular}{@{}p{0.10\textwidth}p{0.16\textwidth}p{0.14\textwidth}ccccccc@{}}
        \toprule
        & \multirow{2}{*}{Description} & Sim-to-real transfer & Multi-agent & Simulator & Knowledge & Learning & Real & \multirow{2}{*}{Application} \\
        &  & and learning details & learning & / Engine & Transfer & Algorithm & Robot/Platform &\\
        \midrule
        \multirow{2}{*}{\textbf{Balaji et al.~\cite{balaji2019deepracer}}} & %
        \scriptsize{DeepRacer: an educational autonomous racing platform.} & %
        \scriptsize{{Random colors and parallel domain randomization}} &        
        \multirow{2}{*}{\shortstack{\cmark (sim only)\\Distr. rollout}} &        
        \multirow{2}{*}{\shortstack{Gazebo\\RoboMaker}} &        
        \multirow{2}{*}{\xmark} &        
        \multirow{2}{*}{\shortstack{PPO}} &        
        \multirow{2}{*}{\shortstack{DeepRacer\\4WD 1:18 Car}} &        
        \multirow{2}{*}{\shortstack{Autonomous \\ racing}} 
        \\
        %
        %
        \multirow{2}{*}{\textbf{Traore et al.~\cite{traore2019continual}}} & %
        \scriptsize{Continual RL with policy distillation and sim-to-real transfer.} & %
        \scriptsize{Continual learning with policy distillation.} &        
        \multirow{2}{*}{\xmark} &        
        \multirow{2}{*}{PyBullet} &        
        \multirow{2}{*}{\shortstack{\cmark Multi-task\\Distillation}} &        
        \multirow{2}{*}{PPO2} &        
        \multirow{2}{*}{\shortstack{Small mobile\\platform}} &        
        \multirow{2}{*}{\shortstack{Robotic \\ navigation}} 
        \\
        \multirow{2}{*}{\textbf{Kaspar et al.~\cite{kaspar2020sim2real}}} & %
        \scriptsize{Sim-to-real transfer for RL without Dynamics Randomization.} & %
        \scriptsize{System identification and a high-quality robot model.} &        
        \multirow{2}{*}{\xmark} &        
        \multirow{2}{*}{PyBullet} &        
        \multirow{2}{*}{\xmark} &        
        \multirow{2}{*}{\shortstack{SAC}} &        
        \multirow{2}{*}{\shortstack{KUKA LBR iiwa\\+WSG50 gripper}} &        
        \multirow{2}{*}{\shortstack{Peg-in-Hole \\ manipulation}} 
        \\
        \multirow{2}{*}{\textbf{Matas et al.~\cite{matas2018sim}}} & %
        \scriptsize{Sim-to-real RL for deformable object manipulation.} & %
        \scriptsize{Stochastic grasping and domain radomization.} &        
        \multirow{2}{*}{\cmark (sim)} &        
        \multirow{2}{*}{PyBullet} &        
        \multirow{2}{*}{\xmark} &        
        \multirow{2}{*}{DDPGfD} &        
        \multirow{2}{*}{\shortstack{7DOF Kinova\\Mico Arm}} &        
        \multirow{2}{*}{\shortstack{Dexterous \\ manipulation}} 
        \\
        \multirow{2}{*}{\textbf{Witman et al.~\cite{witman2019sim}}} & %
        \scriptsize{Sim-to-real RL for thermal effects of an atmospheric pressure plasma jet.} & %
        \scriptsize{Custom physics model and dynamics randomization} &        
        \multirow{2}{*}{\xmark} &        
        \multirow{2}{*}{Custom} &        
        \multirow{2}{*}{\xmark} &        
        \multirow{2}{*}{A3C} &        
        \multirow{2}{*}{\shortstack{kHz-excited\\APPJ in He}} &        
        \multirow{2}{*}{\shortstack{Plasma jet \\ control}} 
        \\
        \multirow{2}{*}{\textbf{Jeong et al.~\cite{jeong2019modelling}}} & %
        \scriptsize{Modeling Generalized Forces with RL for Sim2Real Transfer} & %
        \scriptsize{Modeling and learning state dependent generalized forces.} &        
        \multirow{2}{*}{\xmark} &        
        \multirow{2}{*}{MuJoCo} &        
        \multirow{2}{*}{\xmark} &        
        \multirow{2}{*}{MPO} &        
        \multirow{2}{*}{\shortstack{Rethink Robotics \\ Sawyer}} &        
        \multirow{2}{*}{\shortstack{Nonprehensile \\ manipulation}} 
        \\
        \multirow{2}{*}{\textbf{Arndt et al.~\cite{arndt2019meta}}} & %
        \scriptsize{Meta Reinforcement Learning for Sim2Real Domain Adaptation} & %
        \scriptsize{Domain random. and model-agnostic meta-learning.} &        
        \multirow{2}{*}{\xmark} &        
        \multirow{2}{*}{MuJoCo} &        
        \multirow{2}{*}{\shortstack{\cmark Meta-\\training}} &        
        \multirow{2}{*}{PPO} &        
        \multirow{2}{*}{\shortstack{Kuka LBR \\ 4+ arm}} &        
        \multirow{2}{*}{\shortstack{Manipulation\\(hockey puck)}} 
        \\
        \multirow{2}{*}{\textbf{Breyer et al.~\cite{breyer2018flexible}}} & %
        \scriptsize{Flexible robotic grasping with Sim2Real RL} & %
        \scriptsize{Direct transfer. Elliptic mask to RGB-D images.} &        
        \multirow{2}{*}{\xmark} &        
        \multirow{2}{*}{PyBullet} &        
        \multirow{2}{*}{\xmark} &        
        \multirow{2}{*}{TRPO} &        
        \multirow{2}{*}{\shortstack{ABB YuMi with\\parallel-jaw gripper}} &        
        \multirow{2}{*}{\shortstack{Robotic \\ Grasping}} 
        \\
        \multirow{2}{*}{\textbf{Van Baar et al.~\cite{van2018simulation}}} & %
        \scriptsize{Sim-to-real transfer with robustified policies for robot tasks.} & %
        \scriptsize{Variation of appearance and/ or physics parameters.} &        
        \multirow{2}{*}{\cmark (sim)} &        
        \multirow{2}{*}{\shortstack{MuJoCo\\+Ogre 3D}} &        
        \multirow{2}{*}{\cmark} &        
        \multirow{2}{*}{\shortstack{A3C (sim)\\+Off-policy}} &        
        \multirow{2}{*}{\shortstack{Mitsubishi\\Melfa RV-6SL}} &        
        \multirow{2}{*}{\shortstack{Marble Maze \\ Manipulation}} 
        \\
        \multirow{2}{*}{\textbf{Bassani et al.~\cite{bassani2020learning}}} & %
        \scriptsize{Sim2Real RL for robotic soccer competitions.} & %
        \scriptsize{Domain adaptation and custom simulator for transfer.} &        
        \multirow{2}{*}{\xmark} &        
        \multirow{2}{*}{VSSS-RL} &        
        \multirow{2}{*}{\cmark} &        
        \multirow{2}{*}{\shortstack{DDPG\\/DQN}} &        
        \multirow{2}{*}{VSSS Robot} &        
        \multirow{2}{*}{\shortstack{Robotic \\ Navigation}} 
        \\
        \multirow{2}{*}{\textbf{Qin et al.~\cite{qin2019sim}}} & %
        \scriptsize{Sim2Real for six-legged robots with DRL and curriculum learning.} & %
        \scriptsize{Curriculum learning with inverse kinematics.} &        
        \multirow{2}{*}{\xmark} &        
        \multirow{2}{*}{V-Rep} &        
        \multirow{2}{*}{\cmark} &        
        \multirow{2}{*}{PPO} &        
        \multirow{2}{*}{\shortstack{Six-legged\\robot}} &        
        \multirow{2}{*}{\shortstack{Navigation and\\obstacle avoid.}} 
        \\
        \multirow{2}{*}{\textbf{Vacaro et al.~\cite{vacaro2019sim}}} & %
        \scriptsize{Sim-to-real in reinforcement learning for everyone} & %
        \scriptsize{Domain randomization (light + color + textures).} &        
        \multirow{2}{*}{\cmark (sim)} &        
        \multirow{2}{*}{Unity3D} &        
        \multirow{2}{*}{\xmark} &        
        \multirow{2}{*}{IMPALA} &        
        \multirow{2}{*}{\shortstack{Sainsmart\\robot arm}} &        
        \multirow{2}{*}{\shortstack{Low-cost \\ robot arm}} 
        \\
        \multirow{2}{*}{\textbf{Chaffre et al.~\cite{chaffre2020sim}}} & %
        \scriptsize{Sim-to-Real Transfer with Incremental Environment Complexity} & %
        \scriptsize{SAC training using incremental environment complexity.} &        
        \multirow{2}{*}{\xmark} &        
        \multirow{2}{*}{Gazebo} &        
        \multirow{2}{*}{\xmark} &        
        \multirow{2}{*}{\shortstack{DDPG\\/SAC}} &        
        \multirow{2}{*}{\shortstack{Wifibot\\Lab V4}} &        
        \multirow{2}{*}{\shortstack{Mapless \\ navigation}} 
        \\
        \multirow{2}{*}{\textbf{Kaspar et al.~\cite{kasparreinforcement}}} & %
        \scriptsize{Rl with Cartesian Commands for Peg in Hole Tasks.} & %
        \scriptsize{Dynamics (CMA-ES) and environment randomization.} &        
        \multirow{2}{*}{\xmark} &        
        \multirow{2}{*}{PyBullet} &        
        \multirow{2}{*}{\xmark} &        
        \multirow{2}{*}{SAC} &        
        \multirow{2}{*}{\shortstack{Kuka\\LBR iiwa}} &        
        \multirow{2}{*}{\shortstack{Peg-in-hole \\ tasks}} 
        \\
        \multirow{2}{*}{\textbf{Hundt et al.~\cite{hundt2019good}}} & %
        \scriptsize{Efficient RL for Multi-Step Visual Tasks via Reward Shaping.} & %
        \scriptsize{Direct transfer with custom simulation framework.} &        
        \multirow{2}{*}{\xmark} &        
        \multirow{2}{*}{\shortstack{SPOT\\Framework}} &        
        \multirow{2}{*}{\xmark} &        
        \multirow{2}{*}{\shortstack{SPOT-Q\\+PER}} &        
        \multirow{2}{*}{\shortstack{Universal\\Robot UR5}} &        
        \multirow{2}{*}{\shortstack{Long-term\\multi-step tasks}} 
        \\
        \multirow{2}{*}{\textbf{Pedersen et al.~\cite{pedersen2019sim}}} & %
        \scriptsize{Sim-to-Real Transfer for Gripper Pose Estimation with GAN} & %
        \scriptsize{CycleGANs for domain adaption and transfer.} &        
        \multirow{2}{*}{\xmark} &        
        \multirow{2}{*}{Unity} &        
        \multirow{2}{*}{\xmark} &        
        \multirow{2}{*}{PPO} &        
        \multirow{2}{*}{\shortstack{Panda\\robot}} &        
        \multirow{2}{*}{\shortstack{Robotic \\ Grippers}} 
        \\
        \multirow{2}{*}{\textbf{Ding et al.~\cite{ding2020sim}}} & %
        \scriptsize{Sim-to-Real Transfer for Optical Tactile Sensing} & %
        \scriptsize{Analysis of  different amounts of randomization.} &        
        \multirow{2}{*}{\xmark} &        
        \multirow{2}{*}{PyBullet} &        
        \multirow{2}{*}{\xmark} &        
        \multirow{2}{*}{CNN} &        
        \multirow{2}{*}{\shortstack{Sawyer robot\\+TacTip sensor}} &        
        \multirow{2}{*}{\shortstack{Tactile \\ sensing}} 
        \\
        \multirow{2}{*}{\textbf{Muratore et al.~\cite{muratore2020bayesian}}} & %
        \scriptsize{Data-efficient Bayesian Domain Randomization for sim-to-real} & %
        \scriptsize{Proposed bayesian randomization (BAYR).} &        
        \multirow{2}{*}{\xmark} &        
        \multirow{2}{*}{\shortstack{Custom/\\BoTorch}} &        
        \multirow{2}{*}{\xmark} &        
        \multirow{2}{*}{\shortstack{PPO / RF \\ Classifier}} &        
        \multirow{2}{*}{\shortstack{Quanser\\Qube}} &        
        \multirow{2}{*}{\shortstack{swing-up/\\balancing}} 
        \\
        \multirow{2}{*}{\textbf{Zhao et al.~\cite{wenshuai2020sim2real}}} & %
        \scriptsize{Towards closing the sim-to-real gap in collaborative DRL with perturbances} & %
        \scriptsize{{Domain randomization (custom perturbations)}} &        
        \multirow{2}{*}{\cmark (sim)} &        
        \multirow{2}{*}{Pybullet} &        
        \multirow{2}{*}{\xmark} &        
        \multirow{2}{*}{PPO} &        
        \multirow{2}{*}{\shortstack{Kuka\\(sim-only)}} &        
        \multirow{2}{*}{\shortstack{Robot arm\\reacher}} 
        \\
        \multirow{2}{*}{\textbf{Nachum et al.~\cite{nachum2019multi}}} & %
        \scriptsize{Multi-agent manipulation via locomotion} & %
        \scriptsize{Hierarchichal sim-to-real, model-free, zero-shot transfer.} &        
        \multirow{2}{*}{\cmark} &        
        \multirow{2}{*}{MuJoCo} &        
        \multirow{2}{*}{\xmark} &        
        \multirow{2}{*}{Custom} &        
        \multirow{2}{*}{\shortstack{D’Kitty robo\\(2x)}} &        
        \multirow{2}{*}{\shortstack{Multi-agent\\manipulation}} 
        \\
        \multirow{2}{*}{\textbf{Rajeswaran et al.~\cite{rajeswaran2017learning}}} & %
        \scriptsize{Dexterous manipulation with DRL and demonstrators.} & %
        \scriptsize{Imitation learning via demonstrators with VR.} &        
        \multirow{2}{*}{\xmark} &        
        \multirow{2}{*}{MuJoCo} &        
        \multirow{2}{*}{\xmark} &        
        \multirow{2}{*}{DAPG} &        
        \multirow{2}{*}{\shortstack{ADROIT\\24-DoF Hand}} &        
        \multirow{2}{*}{\shortstack{Multi-fingered\\robot hands}} 
        \\
        \bottomrule
    \end{tabular}
\end{sidewaystable*}


\section{Methodologies for Sim-to-Real Transfer}
\label{sec:methods}

Research in sim-to-real transfer has resulted in an increase of several orders of magnitude in the number of publications over the past few years. Multiple research directions have been followed, and we summarize in this section the most representative methods for sim-to-real transfer.

Table~\ref{tab:papers} lists some of the most relevant and recent works in this field. The most widely used method for learning transfer is domain randomization, with other relevant examples including policy distillation, system identification, or meta-RL. The variability in terms of learning algorithms is higher, with DRL using proximal policy optimization (PPO)~\cite{schulman2017proximal}, trust region policy optimization (TRPO)~\cite{schulman2015trust}, maximum a-posteriori policy optimization (MPO)~\cite{abdolmaleki2018maximum}, asynchronous actor critic (A3C) methods~\cite{mnih2016asynchronous}, soft actor critic (SAC)~\cite{haarnoja2018soft}, or deep deterministic policy gradient (DDPG)~\cite{lillicrap2015continuous}, among others.

\subsection{Zero-shot Transfer}

The most straightforward way of transferring knowledge from simulation to reality is to build a realistic simulator, or to have enough simulated experience, so that the model can be directly applied in real-world settings. This strategy is commonly referred to as zero-shot or direct transfer. System identification to build precise models of the real world and domain randomization are techniques that can be seen as one-shot transfer. 
We discuss both of these separately in Sections~\ref{subsect:identification} and~\ref{subsect:random}.

\subsection{System Identification}\label{subsect:identification}


It is of note that simulators are not faithful representation of the real world. System identification~\cite{kristinsson1992system} is exactly to build a precise mathematical model for a physical system and to make the simulator more realistic careful calibration is necessary. Nonetheless, challenges for obtaining a realistic enough simulator are still existing. For example, it is hard to build high-quality rendered image to simulate the real vision. Furthermore, many physical parameters of the same robot might vary significantly due to temperature, humidity, positioning or its wear-and-tear in time, which brings more difficulty for system identification.

\subsection{Domain Randomization Methods}\label{subsect:random}

\begin{figure*}
    \begin{subfigure}{0.48\textwidth}
        \centering        \includegraphics[width=\textwidth]{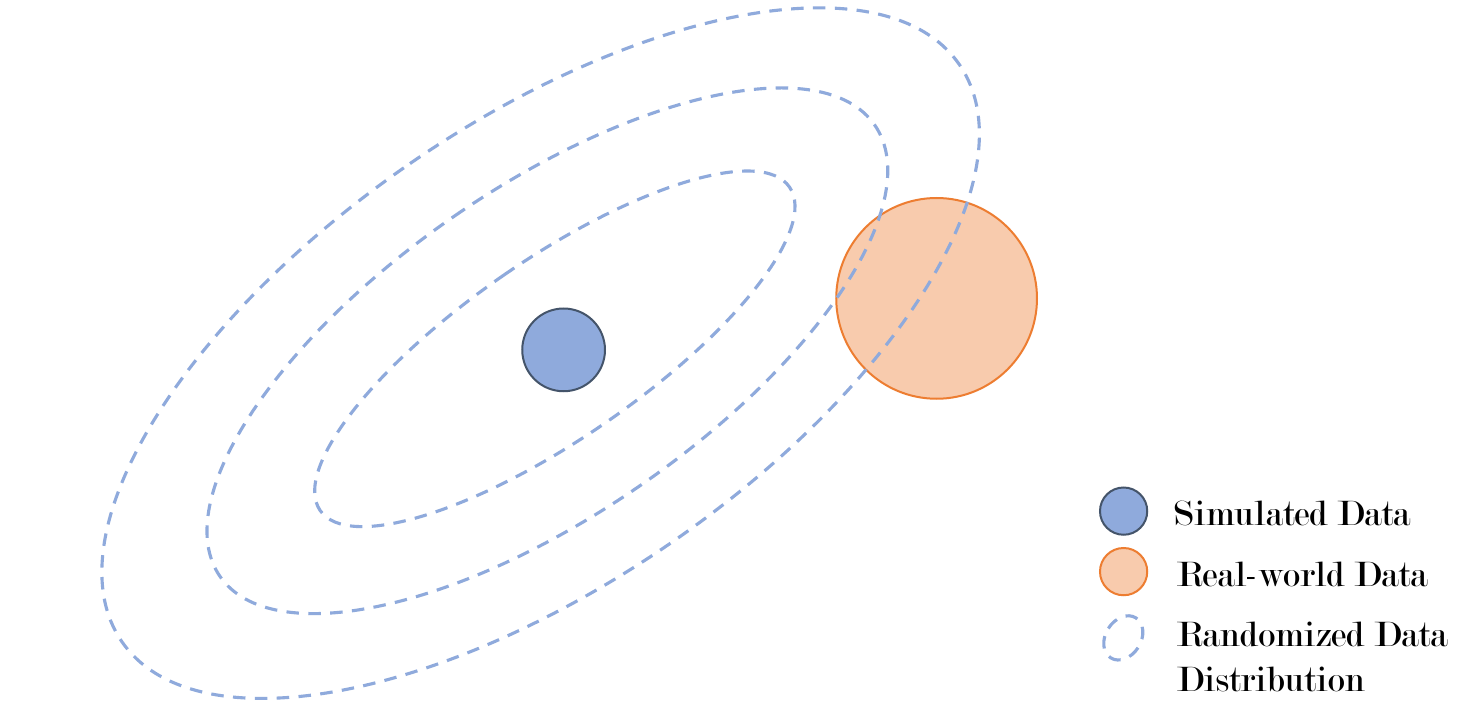}
        \caption{Intuition behind the domain randomization paradigm.}
        \label{fig:DR}
    \end{subfigure}
    \begin{subfigure}{0.48\textwidth}
        \centering
        \includegraphics[width=\textwidth]{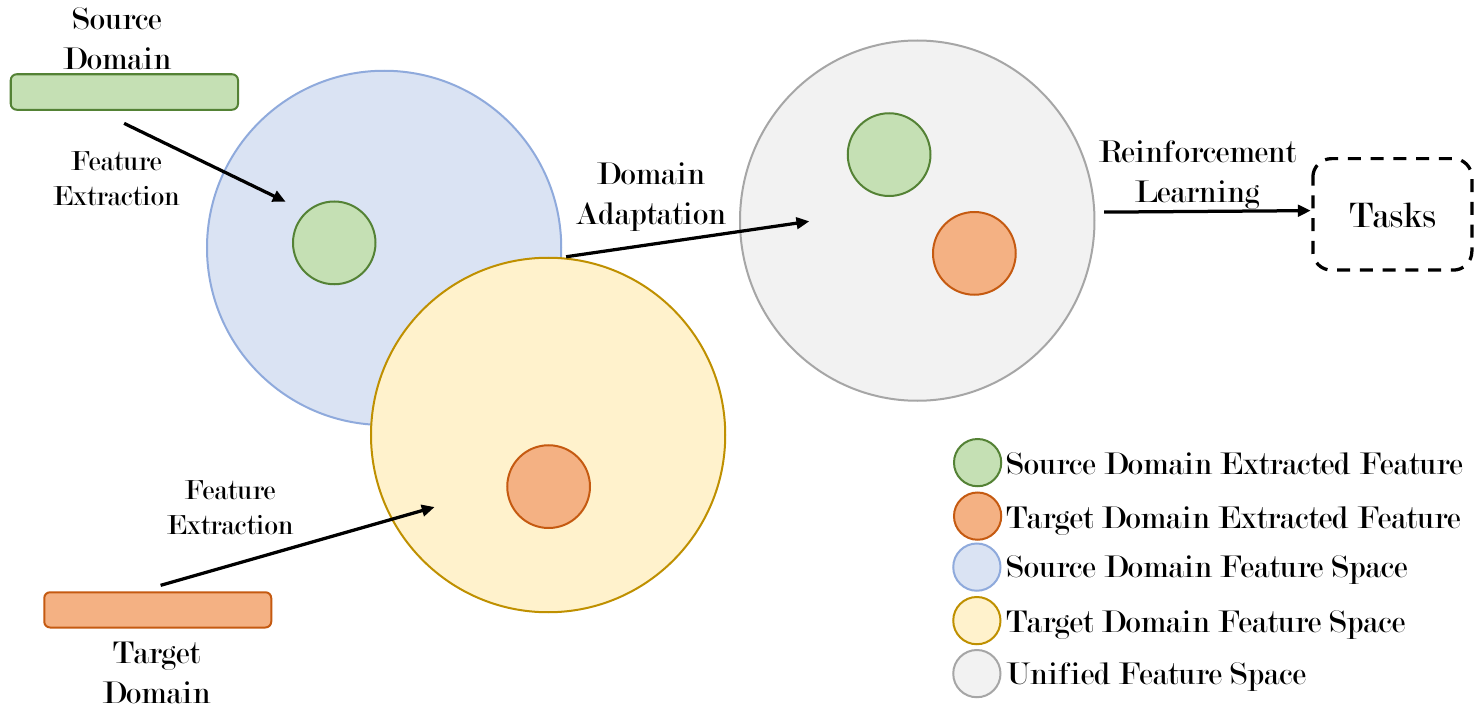}
        \caption{Intuition behind the domain adaptation paradigm.}
        \label{fig:DA}
    \end{subfigure}
    \caption{Illustration of two of the most widely used methods for sim-to-real transfer in DRL. Domain randomization and domain adaptation are often applied as separate techniques, but they can also be applied together.}
\end{figure*}

Domain randomization is the idea that~\cite{tobin2019real}, instead of carefully modeling all the parameters of the real world, we could highly randomize the simulation in order to cover the real distribution of the real-world data despite the bias between the model and real world. Fig.~\ref{fig:DR} shows the paradigm of domain randomization.

According to the components of the simulator randomized, we divide the methods of domain randomization into two kinds: \textit{visual randomization} and \textit{dynamics randomization}. In robotic vision tasks including object localization~\cite{tobin2017domain}, object detection~\cite{tremblay2018training}, pose estimation~\cite{sundermeyer2018implicit}, and semantic segmentation~\cite{yue2019domain}, the training data from simulator always have different textures, lighting, and camera positions from the realistic environments. Therefore, visual domain randomization aims to provide enough simulated variability of the visual parameters at training time such that at test time the model is able to generalize to real-world data. In addition to adding randomization to the visual input, dynamics randomization could also help acquire a robust policy particularly where the controlling policy is needed. To learn dexterous in-hand manipulation policies for a physical five-fingered hand,~\cite{andrychowicz2020learning} randomizes various physical parameters in the simulator, such as object dimensions, objects and robot link masses, surface friction coefficients, robot joint damping coefficients and actuator force gains. Their successful sim-to-real transfer experiments show the powerful effect of domain randomization.

Besides usually making the simulated data randomized to cover the real-world data distribution,~\cite{james2019sim} provides another interesting angle to apply domain randomization. They proposes to translate the randomized simulated image and real-world into the canonical sim images and demonstrate the effectiveness of this sim-to-real approach by training a vision-based closed-loop grasping RL agent in simulation.

\subsection{Domain Adaptation Methods}

Domain adaptation methods use data from source domain to improve the performance of a learned model on a different target domain where data is always less available. Since usually there are different feature spaces between the source domain and target domain, in order to better transfer the knowledge from source data, we should attempt to make these two feature space unified. This is the main spirit of domain adaptation, and can be described by the diagram in Fig.~\ref{fig:DA}.

The research of domain adaptation is broadly conducted recently in vision-based tasks, such as image classification and semantic segmentation~\cite{wang2018deep,hoffman2018cycada}. However, in this paper we focus on the tasks related with reinforcement learning and the ones applied to robotics. In these scenarios, the pure vision related tasks employing domain adaptation play as priors to the succeeding building reinforcement learning agents or other controlling tasks~\cite{james2019sim,bousmalis2018using,yan2017sim}. There is also some image-to-policy work using domain adaptation to generalize the policy learned by synthetic data or speed up the learning on real-world robots~\cite{bousmalis2018using}. Sometimes domain adaptation is used to directly transfer the policy between agents~\cite{gupta2017learning}. 

Specifically, we now formalize the domain adaptation scenarios in a reinforcement learning setting~\cite{higgins2017darla}. Based on the definition of MDP in equation~\eqref{eq:MDP}, we denote the source domain as $D_{S} \equiv (\mathcal{S}_{S}, \mathcal{A}_{S}, \mathcal{P}_{S}, \mathcal{R}_{S})$ and target domain as $D_{T} \equiv (\mathcal{S}_{T}, \mathcal{A}_{T}, \mathcal{P}_{T}, \mathcal{R}_{T})$, respectively. In reinforcement learning scenarios, the states $\mathcal{S}$ of the source and target domain can be quite different $(\mathcal{S}_{S}\neq \mathcal{S}_{T})$ due to the perceptual-reality gap~\cite{rusu2017sim}, while both domains share the action spaces and the transitions $\mathcal{P}$ $(\mathcal{A}_{S} \approx{\mathcal{A}_{T}}, \mathcal{P}_{S} \approx{\mathcal{P}_{T}})$and their reward functions $\mathcal{R}$ have structural similarity $(\mathcal{R}_{S} \approx{\mathcal{R}_{T}})$. 

From the literature, we summarize three common methods for domain adaptation regardless of their tasks. They are \textit{discrepancy-based}, \textit{adversarial-based}, and \textit{reconstruction-based} methods, which can be also used crossly. Discrepancy-based methods measure the feature distance between source and target domain by calculating pre-defined statistical metrics, in order to align their feature spaces~\cite{tzeng2014deep,long2015learning,sun2015return}. Adversarial-based methods build a domain classifier to distinguish whether the features come from source domain or target domain. After being trained, the extractor could produce invariant feature from both source domain and target domain~\cite{ganin2016domain, tzeng2015simultaneous, bousmalis2017unsupervised}. Reconstruction-based methods also aim to find the invariant or shared features between domains. However, they realize this goal by constructing one auxiliary reconstruction task and employ the shared feature to recover the original input~\cite{bousmalis2016domain}. In this way, the shared feature should be invariant and independent with the domains. These three methods provide different angles to make the features from different domains unified, and can be utilized in both vision tasks and RL-based control tasks.

\subsection{Learning with Disturbances}

Domain randomization and dynamics randomization methods focus on introducing perturbations in the simulation environments with the aim of making the agents less susceptible to the mismatches between simulation and reality~\cite{balaji2019deepracer, vacaro2019sim, kasparreinforcement}. The same conceptual idea has been extended in other works, where perturbances has been introduced to obtain more robust agents. For example, in~\cite{wang2020reinforcement}, the authors consider noisy rewards. While not directly related to sim-to-real transfer, noisy rewards can better emulate real-world training of agents. Also, in some of our recent works~\cite{wenshuai2020sim2real, wenshuai2020ubiquitous}, we have considered environmental perturbations that affect differently different agents that are learning in parallel. This is an aspect that needs to be considered when multiple real agents are to be deployed or trained with a common policy.

\subsection{Simulation Environments}

A key aspect in sim-to-real transfer is the choice of simulation. Independently of the techniques utilized for efficiently transferring knowledge to real robots, the more realistic a simulation is the better results that can be expected. The most widely used simulators in the literature are Gazebo~\cite{koenig2004design}, Unity3D, and PyBullet~\cite{coumans2016pybullet} or MuJoCo~\cite{todorov2012mujoco}. Gazebo has the advantage of being widely integrated with the Robot Operating System (ROS) middleware, and therefore can be used together with part of the robotics stack that is present in real robots. PyBullet and MuJoCo, on the other hand, present wider integration with DL and RL libraries and gym environments. In general, Gazebo suits more complex scenarios while PyBullet and MuJoCo provide faster training.

In those cases where system identification for one-shot transfer is the objective, researchers have often built or customized specific simulations that meet problem-specific requirements and constraints~\cite{witman2019sim, bassani2020learning, hundt2019good}.

\section{Application Scenarios}
\label{sec:applications}

Some of the most common applications for DRL in robotics are navigation and dexterous manipulation~\cite{arulkumaran2017brief, kober2013reinforcement}. Owing to the limited operational space in which most robotic arms operate, simulation environments for dexterous manipulation are relatively easier to generate than those for more complex robotic systems. For instance, the Open AI Gym~\cite{brockman2016openai}, one of the most widely used frameworks for reinforcement learning, provides multiple environments for dexterous manipulation. 

\subsection{Dexterous Robotic Manipulation}

Robotic manipulation tasks that have been possible with DRL ranging from learning peg-in-hole tasks~\cite{kasparreinforcement} to deformable object manipulation~\cite{matas2018sim}, and including more dexterous manipulation with multi-fingered hands~\cite{rajeswaran2017learning}, or learning force control policies~\cite{kalakrishnan2011learning}. The latter is particularly relevant for sim-to-real: applying excessive force to real objects might cause damage, while grasping can fail with a lack of force.

In~\cite{matas2018sim}, Matas et al. utilize domain randomization for learning manipulation of deformable objects. The authors identify as one of the main drawbacks of the simulation environment the inability to properly simulate the degree of deformability of the objects, with the real robot being unable to grasp stiffer objects. Moreover, a relevant conclusion from this work is that excessive domain randomization can be detrimental. Specifically, when the number of different colors that were being used for each texture was too large, the performance of the real robot was significantly worse.

\subsection{Robotic Navigation}

While learning navigation with reinforcement learning has been a topic of increasing research interest over the past years~\cite{zhu2017target, zeng2020survey}, the literature focusing on sim-to-real transfer methods is sparse. The first difference with respect to more-established research in learning manipulation is perhaps the lack of standard simulation environments. Owing to the more specific environment and sensor suites that are required for different navigation tasks, custom simulators have often been used~\cite{bassani2020learning, qin2019sim}, or simulation worlds have been created using Unity, Unreal Engine, or Gazebo~\cite{chaffre2020sim, pedersen2019sim}.

Sim-to-real transfer for DRL policies can be applied to complex navigation tasks: from six-legged robots~\cite{qin2019sim} to depth-based mapless navigation~\cite{chaffre2020sim}, including robots for soccer competitions~\cite{bassani2020learning}. In order to achieve a successful transfer to the real world, different methods have been applied in the literature. Of particular interest due to their potential and novelty are the following methods: curriculum learning~\cite{qin2019sim}, incremental environment complexity~\cite{chaffre2020sim}, and continual learning and policy distillation for multiple tasks~\cite{traore2019continual}.

\subsection{Other Applications}

Some other applications of DRL and sim-to-real transfer in robotics that have emerged over the past years are the control of a plasma jet~\cite{witman2019sim}, tactile sensing~\cite{ding2020sim}, or multi-agent manipulation~\cite{nachum2019multi}.

\section{Main Challenges and Future Directions}
\label{sec:discusion}

Albeit the progress presented in the papers we reviewed, sim-to-real remains challenging based on existing methods. For domain randomization, researchers tend to study empirically examining which randomization to add, but it is hard to explain formally how and why it works, which thereby brings the difficulty of designing efficiently simulations and randomization distributions. For domain adaptation, most existing algorithms focus on homogeneous deep domain adaptation, which assumes that the feature spaces between the source and target domains are the same. However, this assumption may not be true in many applications. Thus we expect more exploration to transfer knowledge without this limitation. 

Two of the most promising research directions are: (i) integration of different existing methods for more efficient transfer (e.g., domain randomization and domain adaptation); and (ii) incremental complexity learning, continual learning, and reward shaping for complex or multi-step tasks.

\section{Conclusion}
\label{sec:conclusion}

Reinforcement learning algorithms often rely on simulated data to meet their need for vast amounts of labeled experiences. The mismatch between the simulation environments and real-world scenarios, however, requires further attention to be put to methods for sim-to-real transfer of the knowledge acquired in simulation. This is, to the best of our knowledge, the first survey that focuses on the different approaches being taken for sim-to-real transfer in DRL for robotics.

Domain randomization has been identified as the most widely adopted method for increasing the realism of simulation and better prepare for the real world. However, we have discussed alternative research directions showing promising results. For instance, policy distillation is enabling multi-task learning and more efficient and smaller networks, while meta-learning methods allow for wider variability of tasks.

Multiple challenges remain in this field. While practical implementations show the efficiency of the different methods, wider theoretical and empirical studies are required to better understand the effect of these techniques in the learning process. Moreover, generalization of existing results with a more comprehensive analysis is also lacking in the literature.


\section*{Acknowledgements}

This work was supported by the Academy of Finland's AutoSOS project with grant number 328755.

\bibliographystyle{unsrt}
\bibliography{shortref}

\end{document}